\documentclass[runningheads]{llncs}
\usepackage{graphicx}
\usepackage{amsmath,amssymb} 
\usepackage{color}
\usepackage{multirow}
\usepackage{booktabs}
\usepackage{float}
\usepackage[caption = false]{subfig}
\usepackage{grffile}
\usepackage{gensymb}

\begin{document}
\newcommand{\norm}[1]{\left\lVert #1 \right\rVert}
\newcommand{\TODO}[1]{{\color{red}#1}} 
\title{Occlusion Resistant Object Rotation Regression from Point Cloud Segments} 

\titlerunning{Occlusion Resistant Rotation Regression}
%
\author{Ge Gao \orcidID{0000-0002-8159-9101} \and Mikko Lauri \and Jianwei Zhang \and Simone Frintrop}
%
\authorrunning{G. Gao, M. Lauri, J. Zhang and S. Frintrop}
%

\institute{Department of Informatics, University of Hamburg, Hamburg, Germany 
\email{\{gao,lauri,zhang,frintrop\}@informatik.uni-hamburg.de}}
\maketitle              
\begin{abstract}
Rotation estimation of known rigid objects is important for robotic applications such as dexterous manipulation. 
Most existing methods for rotation estimation use intermediate representations such as templates, global or local feature descriptors, or object coordinates, which require multiple steps in order to infer the object pose.
We propose to directly regress a pose vector from point cloud segments using a convolutional neural network.
Experimental results show that our method can potentially achieve competitive performance compared to a state-of-the-art method, while also showing more robustness against occlusion.
Our method does not require any post processing such as refinement with the iterative closest point algorithm.

\end{abstract}
\section{Introduction} 
\label{sec:introduction}
\begin{figure}[t!]
  \centering
    \includegraphics[width=0.7\textwidth]{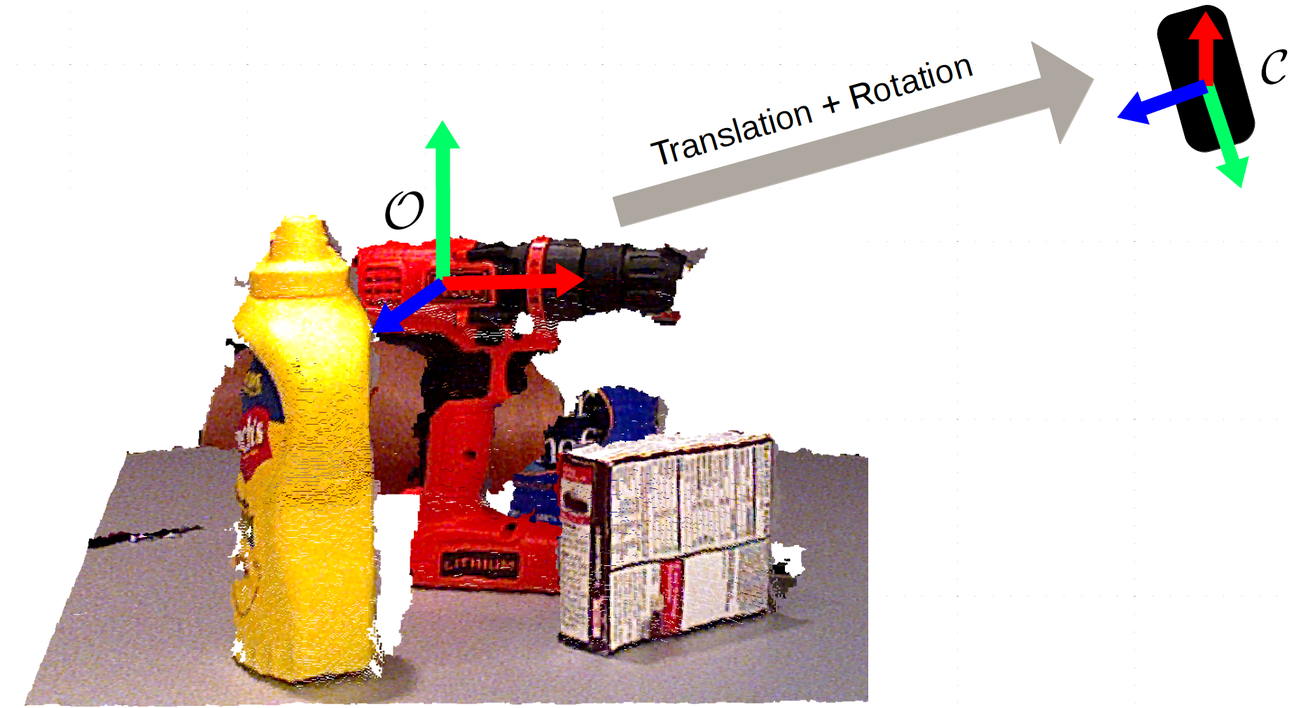}
  \caption{The goal of 6D pose estimation is to find the translation and rotation from the object coordinate frame $\mathcal{O}$ to the camera coordinate frame $\mathcal{C}$.}
\label{fig:prob_state}
\end{figure}

The 6D pose of an object is composed of 3D location and 3D orientation.
The pose describes the transformation from a local coordinate system of the object to a reference coordinate system (e.g. camera or robot coordinate)~\cite{krull2015learning}, as shown in Figure~\ref{fig:prob_state}.
Knowing the accurate 6D pose of an object is necessary for robotic applications such as dexterous grasping and manipulation. 
This problem is challenging due to occlusion, clutter and varying lighting conditions.

Many methods for pose estimation using only color information have been proposed~\cite{kehl2017ssd,rad2017bb8,tekin2018real,Li2018DeepIMDI}.
Since depth cameras are commonly used, there have been many methods using both color and depth information~\cite{brachmann2014learning,kehl2016deep,Jafari2018iPoseI}.
Recently, there are also many CNN based methods~\cite{kehl2016deep,Jafari2018iPoseI}.
In general, methods that use depth information can handle both textured and texture-less objects, and they are more robust to occlusion compared to methods using only color information~\cite{kehl2016deep,Jafari2018iPoseI}.



The 6D pose of an object is an inherently continuous quantity.
Some works discretize the continuous pose space~\cite{Hinterstoisser2011,Hinterstoisser2012accv}, and formulate the problem as classification.
Others avoid discretization by representing the pose using, e.g., quaternions~\cite{xiang2018posecnn}, or the axis-angle representation~\cite{mahendran20173d,do2018deep}.
Work outside the domain of pose estimation has also considered rotation matrices~\cite{qi2017pointnet}, or in a more general case parametric representations of affine transformations~\cite{jaderberg2015spatial}.
In these cases the problem is often formulated as regression.
The choice of rotation representation has a major impact on the performance of the estimation method.


In this work, we propose a deep learning based pose estimation method that uses point clouds as an input.
To the best of our knowledge, this is the first attempt at applying deep learning for directly estimating 3D rotation using point cloud segments.
We formulate the problem of estimating the rotation of a rigid object as regression from a point cloud segment to the axis-angle representation of the rotation.
This representation is constraint-free and thus well-suited for application in supervised learning.

Our experimental results show that our method reaches state-of-the-art performance.
We also show that our method exceeds the state-of-the-art in pose estimation tasks with moderate amounts of occlusion.
Our approach does not require any post-processing, such as pose refinement by the iterative closest point (ICP) algorithm~\cite{chen1992object}.
In practice, we adapt PointNet~\cite{qi2017pointnet} for the rotation regression task. 
Our input is a point cloud with spatial and color information.
We use the geodesic distance between rotations as the loss function.

The remainder of the paper is organized as follows.
Section~\ref{sec:related_work} reviews related work in pose estimation.
In Section~\ref{sec:problem_statement}, we argue why the axis-angle representation is suitable for supervised learning.
We present our system architecture and network details in Section~\ref{sec:system_architecture}.
Section~\ref{sec:experimental_results} presents our experimental results.
In Section~\ref{sec:conclusion} we provide concluding remarks and discuss future work.




\section{Related work} 
\label{sec:related_work}
6D pose estimation using only RGB information has been widely studied~\cite{kehl2017ssd,rad2017bb8,tekin2018real,Li2018DeepIMDI}.
Since this work concentrates on using point cloud inputs, which contain depth information, we mainly review works that also consider depth information.
We also review how depth information can be represented.

\subsection{Pose estimation} 
\label{sub:pose_estimation}

\textbf{RGB-D methods.}
A template matching method which integrates color and depth information is proposed by Hinterstoisser et al.~\cite{Hinterstoisser2011,Hinterstoisser2012accv}.
Templates are built with quantized image gradients on object contour from RGB information and surface normals on object interior from depth information, and annotated with viewpoint information.
The effectiveness of template matching is also shown in~\cite{hodavn2015detection,kehl2015hashmod}.
However, template matching methods are sensitive to occlusions~\cite{kehl2016deep}.

Voting-based methods attempt to infer the pose of an object by accumulating evidence from local or global features of image patches.
One example is the Latent-Class Hough Forest~\cite{tejani2014latent,tejani2018latent} which adapts the template feature from~\cite{Hinterstoisser2011} for generating training data.
During inference stage, a random set of patches is sampled from the input image.
The patches are used in Hough voting to obtain pose hypotheses for verification.

3D object coordinates and object instance probabilities are learned using a Decision Forest in~\cite{brachmann2014learning}.
The 6D pose estimation is then formulated as an energy optimization problem which compares synthetic images rendered with the estimated pose with observed depth values.
3D object coordinates are also used in~\cite{kehl2016deep,Michel2017}.
However, those approaches tend to be very computationally intensive due to generation and verification of  hypotheses~\cite{kehl2016deep}.

Most recent approaches rely on convolutional neural networks (CNNs).
In~\cite{krull2015learning}, the work in~\cite{brachmann2014learning} is extended by adding a CNN to describe the posterior density of an object pose.
A combination of using a CNN for object segmentation and geometry-based pose estimation is proposed in~\cite{jafari2017best}.
PoseCNN \cite{xiang2018posecnn} uses a similar two-stage network, in which the first stage extracts feature maps from RGB input and the second stage uses the generated maps for object segmentation, 3D translation estimation and 3D rotation regression in quaternion format. 
Depth data and ICP are used for pose refinement.
Jafari et al.~\cite{Jafari2018iPoseI} propose a three-stage, instance-aware approach for 6D object pose estimation. 
An instance segmentation network is first applied, followed by an encoder-decoder network which estimates the 3D object coordinates for each segment.
The 6D pose is recovered with a geometric pose optimization step similar to~\cite{brachmann2014learning}.
The approaches~\cite{krull2015learning,Jafari2018iPoseI,xiang2018posecnn} do not directly use CNN to predict the pose.
Instead, they provide segmentation and other intermediate information, which are used to infer the object pose.



\textbf{Point cloud-based.}
Drost et al.~\cite{drost2010model} propose to extract a global model description from oriented point pair features.
With the global description, scene data are matched with models using a voting scheme.
This approach is further improved by~\cite{Hinterstoier2016GoingFW} to be more robust against sensor noise and background clutter.
Compared to~\cite{drost2010model,Hinterstoier2016GoingFW}, our approach uses a CNN to learn the global description.


\subsection{Depth representation} 
\label{sub:depth_representation}
Depth information in deep learning systems can be represented with, e.g., voxel grids~\cite{sedaghat2016orientation,riegler2017octnet}, truncated signed distance functions (TSDF)~\cite{song2016deep}, or point clouds~\cite{qi2017pointnet}.
Voxel grids are simple to generate and use.
Because of their regular grid structure, voxel grids can be directly used as inputs to 3D CNNs.
However, voxel grids are inefficient since they also have to explicitly represent empty space.
They also suffer from discretization artifacts.
TSDF tries to alleviate these problems by storing the shortest distance to the surface represented in each voxel.
This allows a more faithful representation of the 3D information.
In comparison to other depth data representations, a point cloud has a simple representation without redundancy, yet contains rich geometric information.
Recently, PointNet~\cite{qi2017pointnet} has allowed to use raw point clouds directly as an input of a CNN.



\section{Supervised learning for rotation regression} 
\label{sec:problem_statement}
The aim of object pose estimation is to find the translation and rotation that describe the transformation from the object coordinate system $\mathcal{O}$ to the camera coordinate system $\mathcal{C}$ (Figure~\ref{fig:prob_state}).
The translation consists of the displacements along the three coordinate axes, and the rotation specifies the rotation around the three coordinate axes.
Here we concentrate on the problem of estimating rotation.

For supervised learning, we require a loss function that measures the difference between the predicted rotation and the ground truth rotation.
To find a suitable loss function, we begin by considering a suitable representation for a rotation.
We argue that the axis-angle representation is the best suited for a learning task.
We then review the connection of the axis-angle representation to the Lie algebra of rotation matrices.
The Lie algebra provides us with tools needed to define our loss function as the geodesic distance of rotation matrices.
These steps allow our network to directly make predictions in the axis-angle format.

\paragraph{Notation.}
In the following, we denote by $(\cdot)^T$ vector or matrix transpose.
By $\norm{\cdot}_2$, we denote the Euclidean or 2-norm.
We write $\mathrm{I}_{3\times3}$ for the 3-by-3 identity matrix.

\subsection{Axis-angle representation of rotations} 
A rotation can be represented, e.g., as Euler angles, a rotation matrix, a quaternion, or with the axis-angle representation.
Euler angles are known to suffer from gimbal lock discontinuity~\cite{hoag1963apollo}.
Rotation matrices and quaternions have orthogonality and unit norm constraints, respectively.
Such constraints may be problematic in an optimization-based approach such as supervised learning, since they restrict the range of valid predictions.
To avoid these issues, we adopt the axis-angle representation.
In the axis-angle representation, a vector $\mathbf{r}\in\mathbb{R}^3$ represents a rotation of $\theta = \norm{\mathbf{r}}_2$ radians around the unit vector $\frac{\mathbf{r}}{\norm{\mathbf{r}}_2}$~\cite{Hartley2013}.


\subsection{The Lie group $SO(3)$}
The special orthogonal group $SO(3)=\{R \in \mathbb{R}^{3\times 3} \mid RR^T = \mathrm{I}_{3\times 3}, \det R = 1 \}$ is a compact Lie group that contains the 3-by-3 orthogonal matrices with determinant one, i.e., all rotation matrices~\cite{hall2015lie}.
Associated with $SO(3)$ is the Lie algebra $so(3)$, consisting of the set of skew-symmetric 3-by-3 matrices.

Let $\mathbf{r} = \begin{bmatrix}r_1 & r_2 & r_3 \end{bmatrix}^T \in \mathbb{R}^3$ be an axis-angle representation of a rotation. 
The corresponding element of $so(3)$ is the skew-symmetric matrix
\begin{equation}
\label{eq:skewsym}
  \mathbf{r}_{\times} =
 \begin{bmatrix}
  0 & -r_3 & r_2 \\
  r_3 & 0 & -r_1 \\
  -r_2 & r_1 & 0
 \end{bmatrix}.
\end{equation}
The \emph{exponential map} $\exp:so(3)\to SO(3)$ connects the Lie algebra with the Lie group by
\begin{equation}
  \exp(\mathbf{r}_{\times}) = \mathrm{I}_{3\times3} + \frac{\sin\theta}{\theta}\mathbf{r}_{\times} + \frac{1-\cos\theta}{\theta^2} \mathbf{r}^2_{\times},
\end{equation}
where $\theta = \mathbf{r}^T\mathbf{r} = \norm{\mathbf{r}}_2$ as above\footnote{In a practical implementation, the Taylor expansions of $\frac{\sin\theta}{\theta}$ and $\frac{1-\cos\theta}{\theta^2}$ should be used for small $\theta$ for numerical stability.}.

Now let $R$ be a rotation matrix in the Lie group $SO(3)$.
The \emph{logarithmic map} $\log:SO(3) \to so(3)$ connects $R$ with an element in the Lie algebra by
\begin{equation}
  \log(R) = \frac{\phi(R)}{2\sin(\phi(R))}(R-R^T),
\end{equation}
where
\begin{equation}
  \phi(R) = \arccos\left(\frac{\mathrm{trace}(R)-1}{2}\right)
\end{equation}
can be interpreted as the magnitude of rotation related to $R$ in radians.
If desired, we can now obtain an axis-angle representation of $R$ by first extracting from $\log(R)$ the corresponding elements indicated in Eq.~\eqref{eq:skewsym}, and then setting the norm of the resulting vector to $\phi(R)$.

\subsection{Loss function for rotation regression} 
\label{sub:geodesic_distance_on_so}
We regress to a predicted rotation $\hat{\mathbf{r}}$ represented in the axis-angle form.
The prediction is compared against the ground truth rotation $\mathbf{r}$ via a loss function $l:\mathbb{R}^3\times\mathbb{R}^3\to\mathbb{R}_{\geq 0}$.
Let $\hat{R}$ and $R$ denote the two rotation matrices corresponding to $\hat{\mathbf{r}}$ and $\mathbf{r}$, respectively.
We use as loss function the geodesic distance $d(\hat{R}, R)$ of $\hat{R}$ and $R$~\cite{Huynh2009,Hartley2013}, i.e.,
\begin{equation}
  l(\hat{\mathbf{r}}, \mathbf{r}) = d(\hat{R}, R) = \phi(\hat{R}R^T),
\end{equation}
where we first obtain $\hat{R}$ and $R$ via the exponential map, and then calculate $\phi(\hat{R}R^T)$ to obtain the loss value.
This loss function directly measures the magnitude of rotation between $\hat{R}$ and $R$, making it convenient to interpret.
Furthermore, using the axis-angle representation allows to make predictions free of constraints such as the unit norm requirement of quaternions.
This makes the loss function also convenient to implement in a supervised learning approach.


\section{System architecture} 
\label{sec:system_architecture}
\begin{figure}[t!]
  \centering
    \includegraphics[width=.9\textwidth]{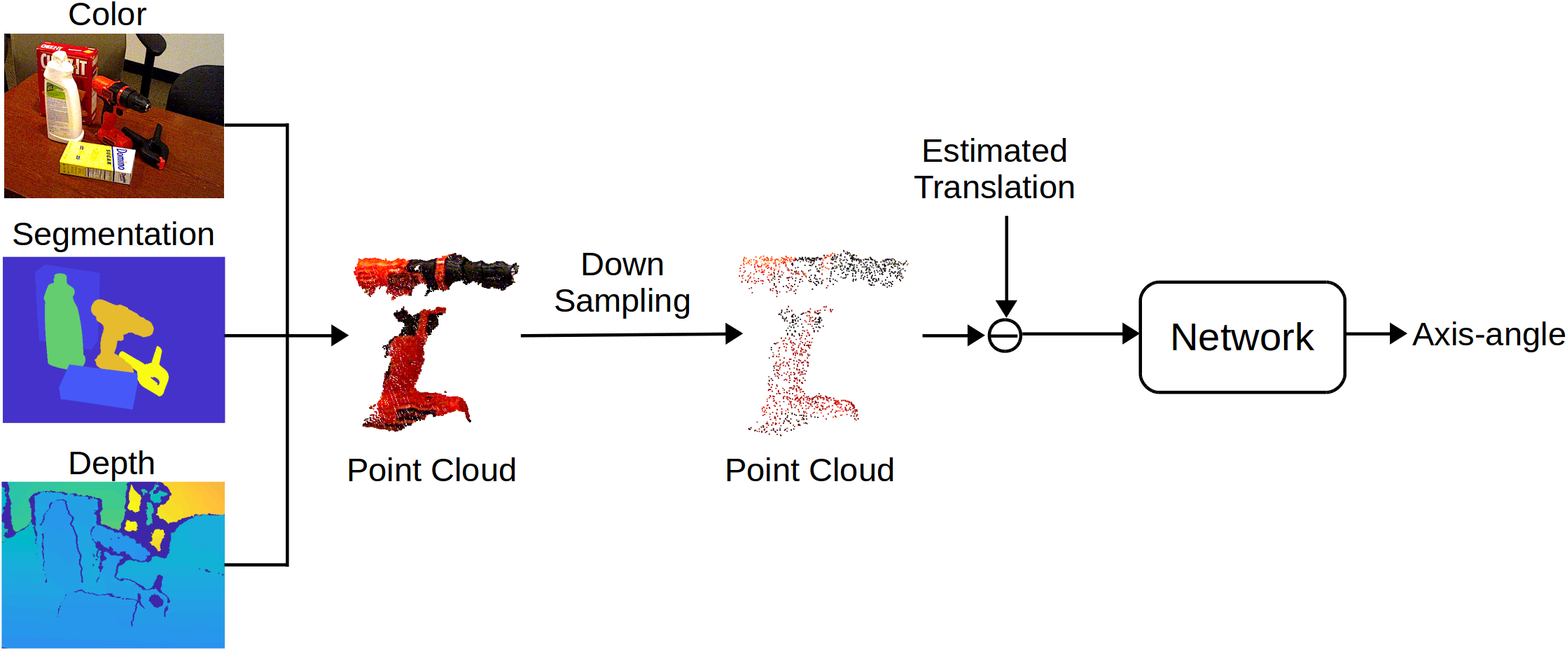}
  \caption{System overview. The color and depth images together with a segmentation of the target object are used to create a point cloud. The segment is randomly downsampled, and the estimated translation of the down-sampled segment is removed. The normalized segment is fed into a network for rotation prediction.}
  \label{fig:system_fig}
\end{figure}

Figure~\ref{fig:system_fig} shows the system overview.
We train our system for a specific target object, in Figure~\ref{fig:system_fig} the drill.
The inputs to our system are the RGB color image, the depth image, and a segmentation mask indicating which pixels belong to the target object.
We first create a point cloud segment of the target object based on the inputs.
Each point has 6 dimensions: 3 dimensions for spatial coordinates and 3 dimensions for color information.
We randomly sample $n$ points from this point cloud segment to create a fixed-size downsampled point cloud.
In all of our experiments, we use $n=256$.
We then remove the estimated translation from the point coordinates to normalize the data.
The normalized point cloud segment is then fed into a network which outputs a rotation prediction in the axis-angle format.
During training, we use the ground truth segmentation and translation.
As we focus on the rotation estimation, during testing, we apply the segmentation and translation outputs of PoseCNN~\cite{xiang2018posecnn}.

We consider two variants for our network presented in the following subsections.
The first variant processes the point cloud as a set of independent points without regard to the local neighbourhoods of points.
The second variant explicitly takes into account the local neighbourhoods of a point by considering its nearest neighbours.

\subsection{PointNet (PN)} 
\label{sub:pointnet_}
\begin{figure}[t]
  \centering
    \includegraphics[width=.8\textwidth]{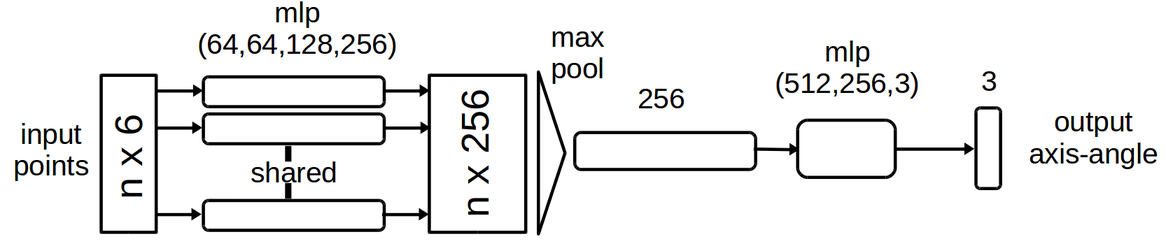}
  \caption{Network architecture. The numbers in parentheses indicate number of MLP layers, and numbers not in parentheses indicate intermediate vector dimensionality. A feature vector for each point is learned using shared weights. A max pooling layer then aggregates the individual features into a global feature vector. Finally, a regression network with 3 fully-connected layers outputs the rotation prediction.}
  \label{fig:network_fig}
\end{figure}

Our PN network is based on PointNet~\cite{qi2017pointnet}, as illustrated in Figure~\ref{fig:network_fig}.
The PointNet architecture is invariant to all $n!$ possible permutations of the input point cloud, and hence an ideal structure for processing raw point clouds.
The invariance is achieved by processing all points independently using multi-layer perceptrons (MLPs) with shared weights.
The obtained feature vectors are finally max-pooled to create a global feature representation of the input point cloud.
Finally, we attach a three-layer regression MLP on top of this global feature to predict the rotation.

\subsection{Dynamic nearest neighbour graph (DG)} 
\label{sub:dynamic_nearest_neighbour_graph}
In the PN architecture, all features are extracted based only on a single point.
Hence it does not explicitly consider the local neighbourhoods of individual points.
However, local neighbourhoods can contain useful geometric information for pose estimation~\cite{rusu2010fast}.
The local neighbourhoods are considered by an alternative network structure based on the dynamic nearest-neighbour graph network proposed in~\cite{wang2018dynamic}.
For each point $P_i$ in the point set, a $k$-nearest neighbor graph is calculated.
In all our experiments, we use $k=10$.
The graph contains directed edges $(i,j_{i1}),\dots,(i,j_{ik})$, such that $P_{j_{i1}},\dots,P_{j_{ik}}$ are the $k$ closest points to $P_i$.
For an edge $e_{ij}$, an edge feature $\begin{bmatrix}P_i, & (P_j - P_i) \end{bmatrix}^T$ is calculated.
The edge features are then processed in a similar manner as in PointNet to preserve permutation invariance.
This dynamic graph convolution can then be repeated, now calculating the nearest neighbour graph for the feature vectors of the first shared MLP layer, and so on for the subsequent layers.
We use the implementation\footnote{https://github.com/WangYueFt/dgcnn} provided by authors from~\cite{wang2018dynamic}, and call the resulting network DG for dynamic graph.

\section{Experimental results} 
\label{sec:experimental_results}
This section shows experimental results of the proposed approach on the YCB video dataset~\cite{xiang2018posecnn}, and compares the performance with state-of-the-art PoseCNN method~\cite{xiang2018posecnn}.
Besides prediction accuracy, we investigate the effect of occlusions and the quality of the segmentation and translation estimates.

\subsection{Experiment setup} 
\label{sub:experiment_setup}
The YCB video dataset~\cite{xiang2018posecnn} is used for training and testing with the original train/test split.
The dataset contains 133,827 frames of 21 objects selected from the YCB object set~\cite{calli2015benchmarking} with 6D pose annotation.
80,000 frames of synthetic data are also provided as an extension to the training set.

\begin{figure}[t]
  \centering
    \includegraphics[width=.5\textwidth]{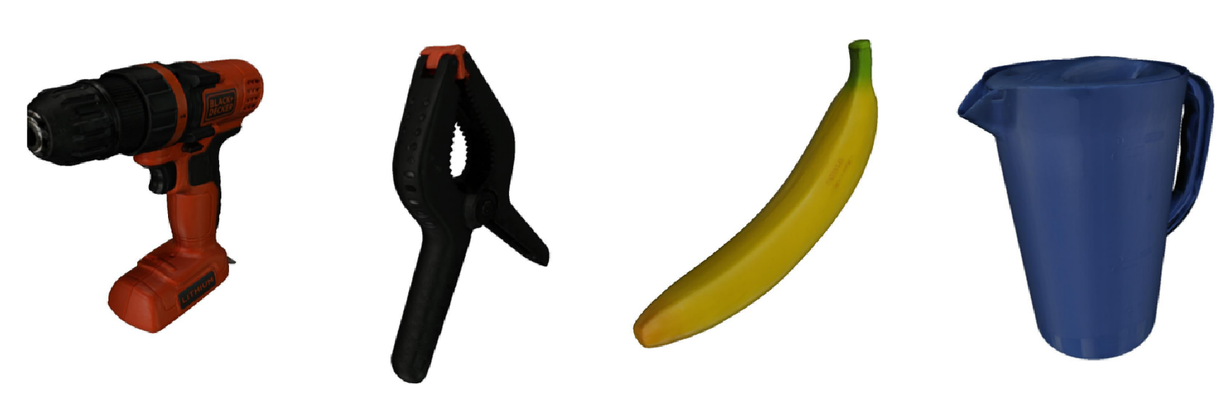}
  \caption{Testing objects. From left to right: power drill, extra large clamp, banana, pitcher base.}
  \label{fig:test_objects}
\end{figure}

We select a set of four objects to test on, shown in Figure \ref{fig:test_objects}.
As our approach does not consider object symmetry, we use objects that have 1-fold rotational symmetry (power drill, banana and pitcher base) or 2-fold rotational symmetry (extra large clamp). 

We run all experiments using both the PointNet based (PN) and dynamic graph (DG) networks.
During training, Adam optimizer is used with learning rate $0.008$, and batch size of $128$.
Batch normalization is applied to all layers.
No dropout is used.

For training, ground truth segmentations and translations are used as the corresponding inputs shown in Fig.~\ref{fig:system_fig}.
While evaluating 3D rotation estimation in Subsection~\ref{sub:rotation_estimation_results}, the translation and segmentation predicted by PoseCNN are used.

We observed that the color information represented by RGB color space varies in an inconsistent manner across different video sequences, hence all the following experimental results are obtained only with XYZ coordinate information of point cloud. Moreover, our current system does not deal with classification problem, individual network is trained for each object. Due to the difference of experimental setup between our method and PoseCNN, the performance comparison are mainly for illustrating the potential of proposed approach.


\subsection{Evaluation metrics} 
\label{sub:evaluation_metrics}
For evaluating rotation estimation, we directly use geodesic distance described in Section~\ref{sec:problem_statement} to quantify the rotation error.
We evaluate 6D pose estimation using average distance of model points (ADD) proposed in~\cite{Hinterstoisser2012accv}.
For a 3D model $\mathcal{M}$ represented as a set of points, with ground truth rotation $R$ and translation $\mathbf{t}$, and estimated rotation $\hat{R}$ and translation $\mathbf{\hat{t}}$, the ADD is defined as:
\begin{align}
\mathrm{ADD}=\frac{1}{m}\displaystyle\sum_{\mathbf{x}\in\mathcal{M}} \norm{ (R\mathbf{x}+\mathbf{t})-(\hat{R}\mathbf{x}+\hat{\mathbf{t}}) }_2,
\end{align}
where $m$ is the number of points.
The 6D pose estimate is considered to be correct if ADD is smaller than a given threshold.


\subsection{Rotation estimation} 
\label{sub:rotation_estimation_results}

\begin{figure}[t]
\subfloat{\includegraphics[width = .5\textwidth]{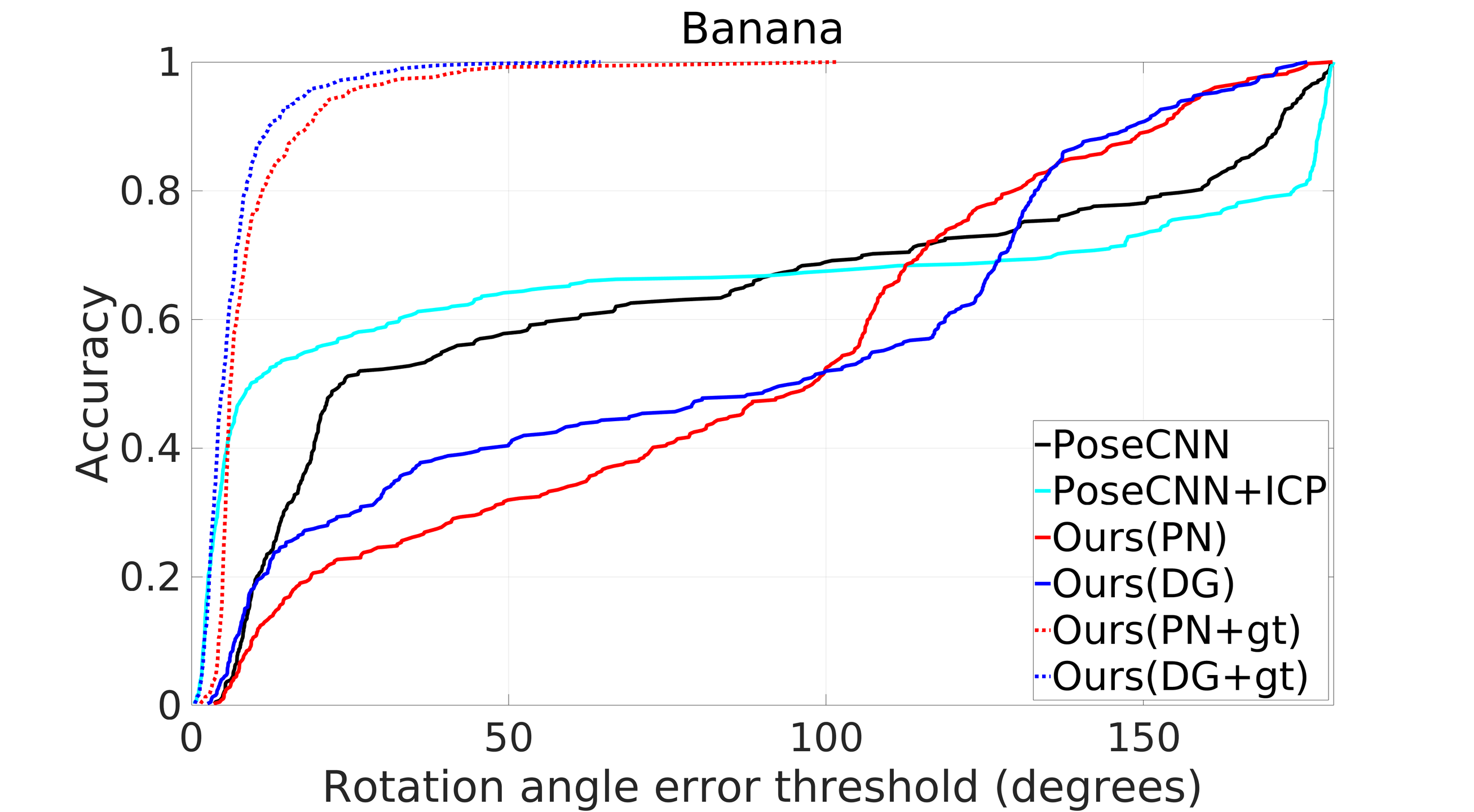}} 
\subfloat{\includegraphics[width = .5\textwidth]{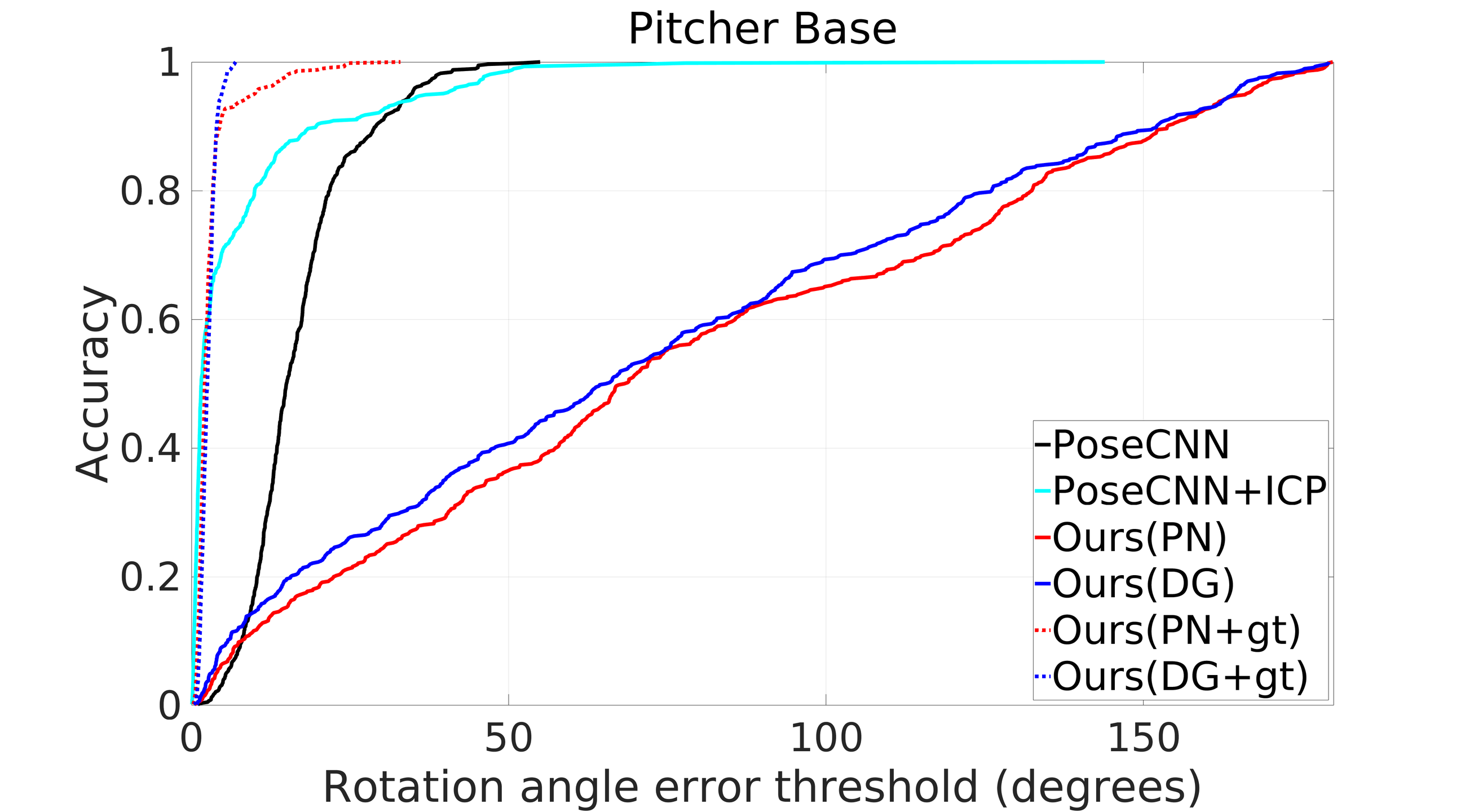}}\\
\subfloat{\includegraphics[width = .5\textwidth]{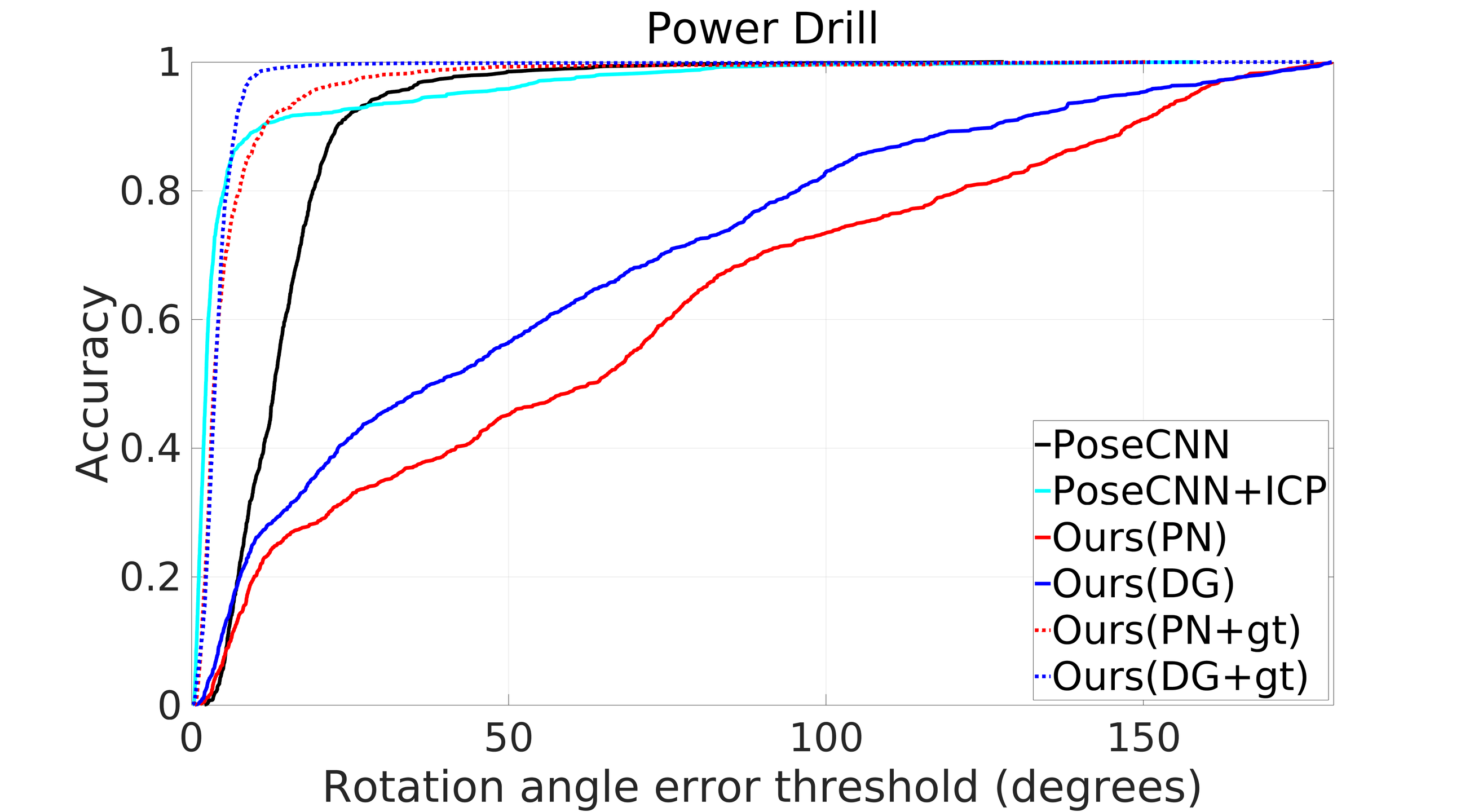}}
\subfloat{\includegraphics[width = .5\textwidth]{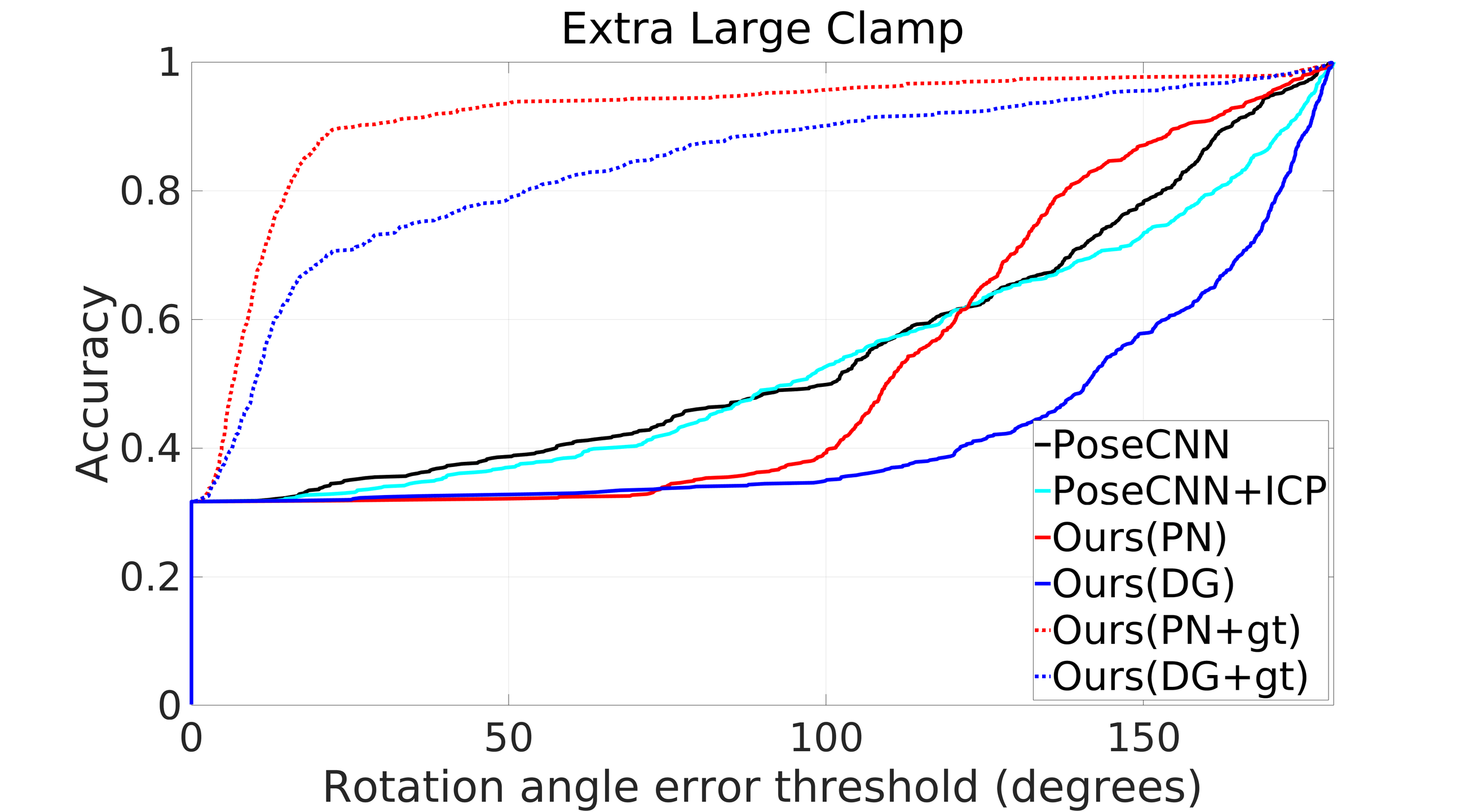}} 
\caption{Accuracy of rotation angle prediction shows the fraction of predictions with error smaller than the threshold. Results are shown for our method and PoseCNN~\cite{xiang2018posecnn}. The additional +gt denotes the variants where ground truth segmentation is provided.}
\label{fig:result_geo_dist}
\end{figure}

Figure~\ref{fig:result_geo_dist} shows the estimation accuracy as function of the rotation angle error threshold, i.e., the fraction of predictions that have an angle error smaller than the horizontal axis value.
Results are shown for PoseCNN, PoseCNN with ICP refinement (PoseCNN+ICP), and our method with PointNet structure (PN), and with dynamic graph structure (DG).
To determine the effect of the translation and segmentation input, we additionally test our methods while giving the ground truth translation and segmentation as input.
The cases with ground truths provided are indicated by +gt, and shown with a dashed line.

The performance without ground truth translation and segmentation is significantly worse than the performance with ground truth information. 
This shows that good translation and segmentation results are crucial for accurate rotation estimation.
Also, by using ground truth information, the performance for extra large clamp (2-fold rotational symmetry) is worse than other objects, which illustrates that the object symmetry should be taken into consideration during learning process.

The results also confirm the fact that ICP based refinement usually only improves the estimation quality if the initial guess is already good enough. 
When the initial estimation is not accurate enough, the use of ICP can even decrease the accuracy, as shown by the PoseCNN+ICP curve falling below the PoseCNN curve for large angle thresholds.

\textbf{Effect of occlusion.}
We quantify the effect of occlusion on the rotation prediction accuracy.
For a given frame and target object, we estimate the occlusion factor $O$ of the object by
\begin{align}
O = 1 - \frac{\lambda}{\mu},
\end{align}
where $\lambda$ is the number of pixels in the 2D ground truth segmentation, and $\mu$ is the number of pixels in the projection of the 3D model of the object onto the image plane using the camera intrinsic parameters and the ground truth 6D pose, when we assume that the object would be fully visible.
We noted that for the test frames of the YCB-video dataset $O$ is mostly below 0.5.
We categorize $O < 0.2$ as low occlusion and $O\geq 0.2$ as moderate occlusion.

\begin{table}[t]
\centering
\caption{Average rotation angle error in degrees with $95\%$ confidence interval in frames with low ($O<0.2$) and moderate (mod, $O\geq 0.2$) occlusion}
\begin{tabular}{@{}c@{}c@{}c@{}c@{}c@{}c@{}c@{}}
\toprule
Object      & \multicolumn{2}{c}{Banana} & \multicolumn{2}{c}{Power Drill} & \multicolumn{2}{c}{Extra Large Clamp} \\ \cmidrule(lr){1-1} \cmidrule(lr){2-3} \cmidrule(lr){4-5} \cmidrule(lr){6-7}
Occlusion   & low              & mod              & low              & mod             & low              & mod         \\ \midrule
PoseCNN \cite{xiang2018posecnn}       & 62.0\degree$\pm$3.1\degree   & 8.2\degree$\pm$0.25\degree    & 14.7\degree$\pm$0.3\degree   & 37.4\degree$\pm$2.4\degree  & 109.8\degree$\pm$2.0\degree  & 151.0\degree$\pm$3.6\degree \\ 
PoseCNN+ICP    & 56.5\degree$\pm$3.4\degree   & 7.1\degree$\pm$0.9\degree     & 6.9\degree$\pm$0.4\degree     & 44.1\degree$\pm$3.5\degree  & 115.5\degree$\pm$2.0\degree  & 140.5\degree$\pm$6.0\degree \\ 
Ours (PN)     & 93.3\degree$\pm$2.2\degree    & 107.4\degree$\pm$1.5\degree   & 65.1\degree$\pm$1.3\degree   & $94.7\degree\pm6.1\degree$   & 115.5\degree$\pm$1.4\degree  & 138.4\degree$\pm$4.3\degree \\ 
Ours (DG)    & 82\degree$\pm$2.5\degree    & 130.4\degree$\pm$1.5\degree    & 51.3\degree$\pm$1.2\degree    & 130.5\degree$\pm$4.1\degree    & 145.7\degree$\pm$1.7\degree   & 134.2\degree$\pm$3.1\degree   \\ \midrule
Ours (PN+gt) & 9.9\degree$\pm$0.5\degree    & $\mathbf{5.7\degree\pm0.1\degree}$   & 6.5\degree$\pm$0.3\degree    & 13\degree$\pm$0.8\degree   & $\mathbf{11.2\degree\pm0.4\degree}$    & $\mathbf{5.7\degree\pm0.4\degree}$  \\ 
Ours (DG+gt) & $\mathbf{7.1\degree\pm0.3\degree}$    & 9.8\degree$\pm$1.2\degree    & $\mathbf{4.3\degree\pm0.2\degree}$    & $\mathbf{2.6\degree\pm0.3\degree}$   & 34.1\degree$\pm$1.6\degree   &  68.2\degree$\pm$8.9\degree   \\ \bottomrule
\end{tabular}
\label{tab:occlusion}
\end{table}

Table \ref{tab:occlusion} shows the average rotation angle error (in degrees) and its $95\%$ confidence interval\footnote{The results for pitcher base are not reported here since all samples in testing set for pitcher base have low occlusion.} for PoseCNN and our method in the low and moderate occlusion categories.
We also investigated the effect of the translation and segmentation by considering variants of our methods that were provided with the ground truth translation and segmentation.
These variants are shown in the table indicated by +gt.

We observe that with ground truth information, our methods shows potential in cases of both low and moderate occlusion.
Furthermore, with the dynamic graph architecture (DG), the average error tends to be lower for 1-fold rotational symmetry objects.
This shows the local neighbourhood information extracted by DG is useful for rotation estimation when there is no pose ambiguity.
One observation is that for banana, the rotation error in low occlusion is significantly higher than it is in the moderate case for PoseCNN.
This is because near $25\%$ of the test frames in low occlusion case present an rotation error in range of $160\degree$ to $180\degree$.

\begin{figure*}[t]
\centering
\begin{tabular}{ccccc}
& Ground Truth &PoseCNN+ICP & Ours (DG) & Ours (DG+gt) \\
$O=0.3$&
\subfloat{\includegraphics[width = .2\textwidth]{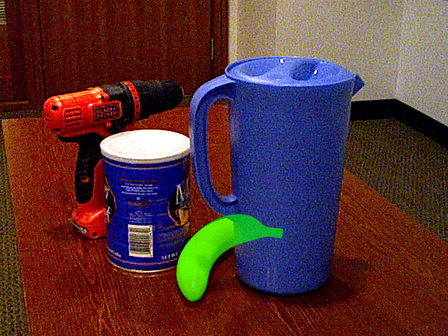}} &
\subfloat{\includegraphics[width = .2\textwidth]{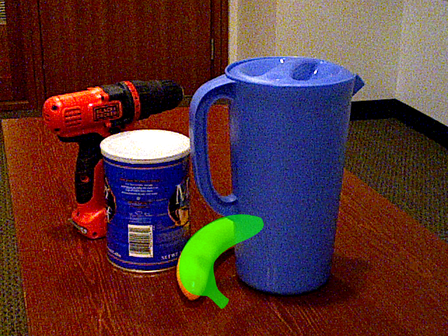}} &
\subfloat{\includegraphics[width = .2\textwidth]{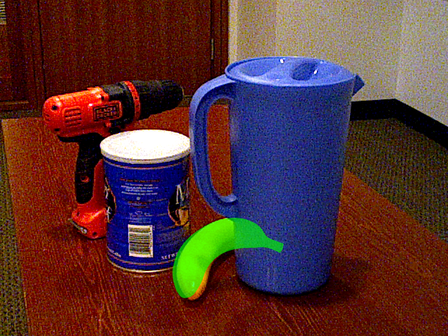}} &
\subfloat{\includegraphics[width = .2\textwidth]{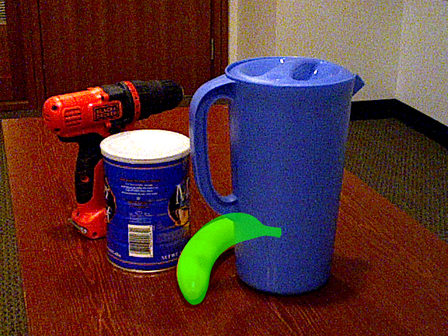}}\\
$O=0.45$&
\subfloat{\includegraphics[width = .2\textwidth]{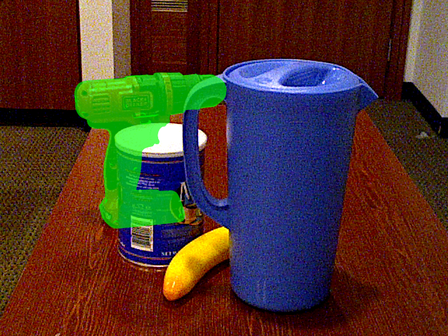}} &
\subfloat{\includegraphics[width = .2\textwidth]{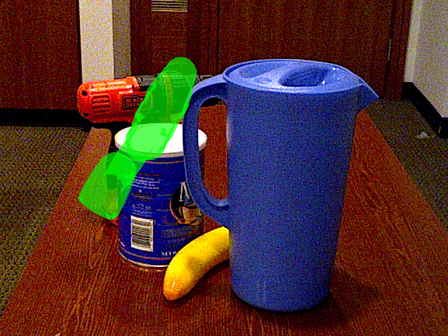}}&
\subfloat{\includegraphics[width = .2\textwidth]{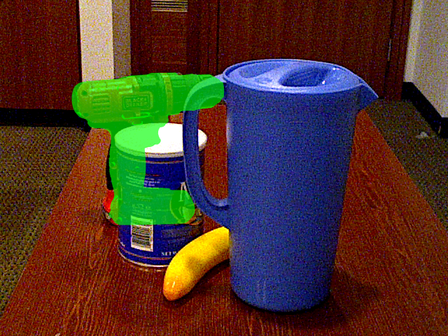}}&
\subfloat{\includegraphics[width = .2\textwidth]{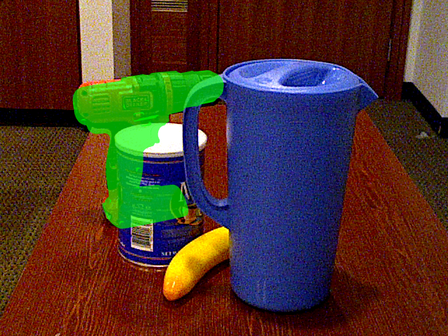}}\\
$O=0.02$& 
\subfloat{\includegraphics[width = .2\textwidth]{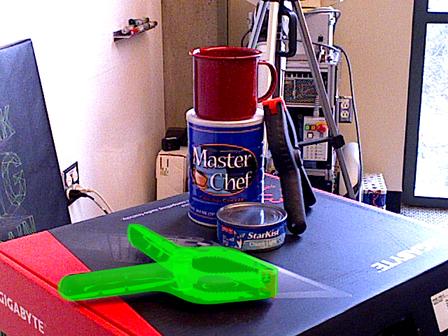}} &
\subfloat{\includegraphics[width = .2\textwidth]{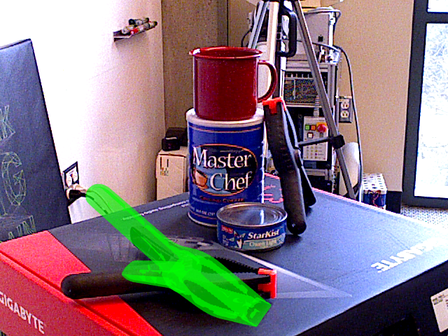}}&
\subfloat{\includegraphics[width = .2\textwidth]{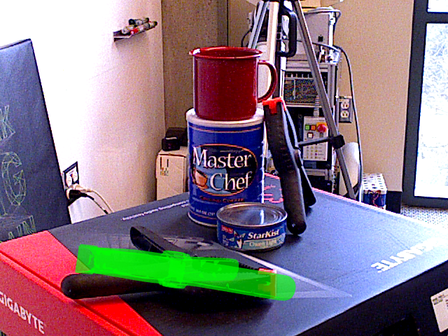}}&
\subfloat{\includegraphics[width = .2\textwidth]{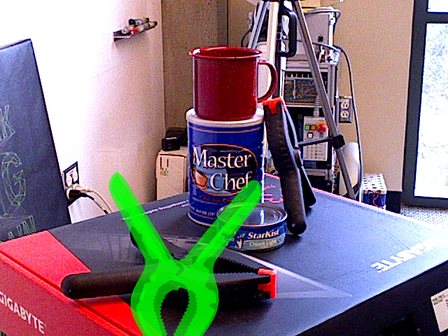}} \\
$O=0.15$&
\subfloat{\includegraphics[width = .2\textwidth]{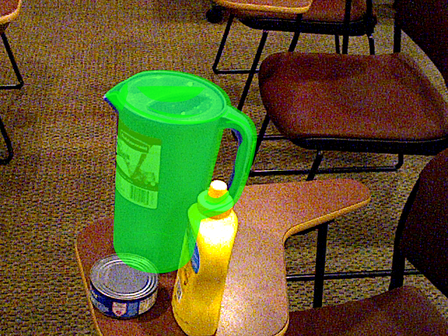}} &
\subfloat{\includegraphics[width = .2\textwidth]{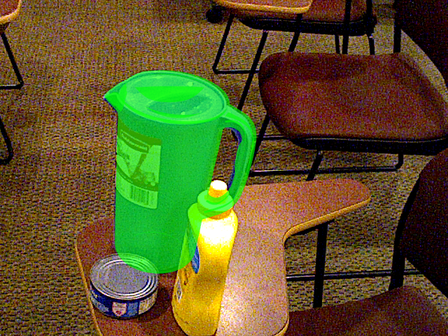}} &
\subfloat{\includegraphics[width = .2\textwidth]{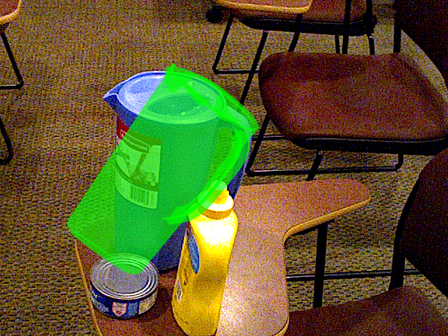}} &
\subfloat{\includegraphics[width = .2\textwidth]{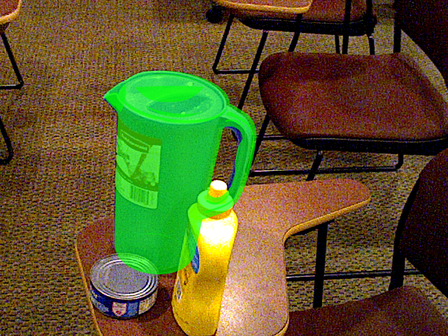}}
\end{tabular}
\caption{Qualitative results for rotation estimation. The number on the left indicates the occlusion factor $O$ for the target object. Then, from left to right: ground truth, PoseCNN~\cite{xiang2018posecnn} with ICP refinement, our method using dynamic graph (DG) with PoseCNN segmentation, and dynamic graph with ground truth segmentation (DG+gt). The green overlay indicates the ground truth pose, or respectively, the predicted pose of the target object. Ground truth translation is used in all cases.}
\label{fig:result_visual_rot}
\end{figure*}

Qualitative results for rotation estimation are shown in Figure~\ref{fig:result_visual_rot}.
In the leftmost column, the occlusion factor $O$ of the target object is denoted.
Then, from left to right, we show the ground truth, PoseCNN+ICP, and our method using DG and our method using DG with ground truth translation and segmentation (DG+gt) results.
In all cases, the ground truth pose, or respectively, the pose estimate, are indicated by the green overlay on the figures.
To focus on the difference in the rotation estimate, we use the ground truth translation for all methods for the visualization.
The rotation predictions for Ours (DG) are still based on translation and segmentation from PoseCNN.

The first two rows of Figure~\ref{fig:result_visual_rot} show cases with moderate occlusion.
When the discriminative part of the banana is occluded (top row), PoseCNN can not recover the rotation, while our method still produces a good estimate.
The situation is similar in the second row for the drill.
The third row illustrates that the quality of segmentation has a strong impact on the accuracy of rotation estimation.
In this case the segmentation fails to detect the black clamp on the black background, which leads to a poor rotation estimate for both PoseCNN and our method.
When we provide the ground truth segmentation (third row, last column), our method is still unable to recover the correct rotation due to the pose ambiguity.

\section{Conclusion} 
\label{sec:conclusion}
We propose to directly predict the 3D rotation of a known rigid object from a point cloud segment.
We use axis-angle representation of rotations as the regression target.
Our network learns a global representation either from individual input points, or from point sets of nearest neighbors.
Geodesic distance is used as the loss function to supervise the learning process.
Without using ICP refinement, experiments shows that the proposed method can reach competitive and sometimes superior performance compared to PoseCNN.

Our results show that point cloud segments contain enough information for inferring object pose.
The axis-angle representation does not have any constraints, making it a suitable regression target.
Using Lie algebra as a tool provides a valid distance measure for rotations.
This distance measure can be used as a loss function during training.

We discovered that the performance of our method is strongly affected by the quality of the target object translation and segmentation, which will be further investigated in future work.
We will extend the proposed method to full 6D pose estimation by additionally predicting the object translations.
We also plan to integrate object classification into our system, and study a wider range of target objects.

\section*{Acknowledgments}
This work was partially funded by the German Science Foundation (DFG) in project Crossmodal Learning, TRR 169.

\clearpage
%
%
%
%
\bibliographystyle{splncs04}
\bibliography{egbib}
\end{document}